\documentclass[lettersize,journal]{IEEEtran}
\usepackage{amsmath,amsfonts}
\usepackage{algorithmic}
\usepackage{algorithm}
\usepackage{array}
\usepackage[caption=false,font=normalsize,labelfont=sf,textfont=sf]{subfig}
\usepackage{textcomp}
\usepackage{stfloats}
\usepackage{url}
\usepackage{verbatim}
\usepackage{graphicx}
\usepackage{booktabs}
\usepackage{hyperref}
\usepackage{cite}
\begin{document}

\title{Bayesian Network Modeling of Causal Influence within Cognitive Domains and \\ Clinical Dementia Severity Ratings for \\ Western and Indian Cohorts}

\author{Wupadrasta Santosh Kumar, Sayali Rajendra Bhutare, Neelam Sinha, Thomas Gregor Issac
\thanks{Manuscript updated on Aug 14, 2024.}}




\maketitle

\begin{abstract}
This study investigates the causal relationships between Clinical Dementia Ratings (CDR) and its six domain scores across two distinct aging datasets: the Alzheimer’s Disease Neuroimaging Initiative (ADNI) and the Longitudinal Aging Study of India (LASI). Using Directed Acyclic Graphs (DAGs) derived from Bayesian network models, we analyze the dependencies among domain scores and their influence on the global CDR. Our approach leverages the PC algorithm to estimate the DAG structures for both datasets, revealing notable differences in causal relationships and edge strengths between the Western and Indian populations. The analysis highlights a stronger dependency of CDR scores on memory functions in both datasets, but with significant variations in edge strengths and node degrees. By contrasting these findings, we aim to elucidate population-specific differences and similarities in dementia progression, providing insights that could inform targeted interventions and improve understanding of dementia across diverse demographic contexts.
\end{abstract}


\section{Introduction}
\IEEEPARstart{T}{he} cognitive assessment of a healthy individual has been an important field of investigation for decades. Understanding if the subject under examination is on the verge of developing a cognitive disorder remains a vital challenge for the scientific community. Dementia is a chronic brain syndrome which deranges various higher cognitive functions such as memory, thinking, orientation, comprehension, calculation, language and judgement, ultimately interfering with their daily functioning. Especially in the case of dementia, the prodromal symptoms are difficult to study, even in population studies with follow-up of more than 8 to 10 years. Further, studies state that the prodromal symptoms of dementia might differ, based on the onset age, leading to different symptom profiles. For example, older subjects had decreased loads of AD pathology \cite{berg1998clinicopathologic}. While AD pathology increased with age in elderly non-demented individuals \cite{neuropathology2001pathological}. Nevertheless, the influence of age on prodromal symptoms of dementia has not been examined in long-term prospective studies \cite{sacuiu2009pattern}. Besides, most of the identifiable symptoms, such as impaired memory function, develop post onset. Interestingly, studies have also reported that the dementia onset either led to abrupt decline in the memory function \cite{backman2001stability,amieva20059} or consequently gave rise to other behavioral changes, such as language and executive functions \cite{sacuiu2005prodromal}. In addition, memory impairment alone did not indicate dementia onset \cite{saxton2004preclinical}. Conclusively, the challenge of identifying dementia onset, came to be handled using physical examination, cognitive tests, and neuro-imaging investigations \cite{brown2015use}. Of these, the cognitive tests were primarily considered to confirm if the cognitive decline is inexplicable by any other mental disorders. Such tests examine not just the decline in cognitive memory, but also various psychometric changes, related to intellectual capabilities and personality traits, allowing clinicians to check if the memory impairment in a subject is a cause of concern or if it requires further investigation. In dementia examination, there are various neuro-psychological test batteries that are used, such as Addenbrooke’s Cognitive Examination-Revised (ACE-R) as well as global assessment tests such as Clinical Dementia Rating (CDR) \cite{hatfield2009diagnostic}.

Clinical Dementia Rating (CDR) \cite{morris1993clinical} is one of the widely used global assessment scores for detection and staging of dementia that indicates not just the decline in cognitive function or specific symptoms but also the impact of dementia on the lives of patients. CDR is also highly useful in studying the impact of treatments or interventions. The clinical dementia rating (CDR) is based on six domains, namely: memory, orientation, judgement (problem solving and financial affairs), community affairs (function at work, volunteering and social groups), home (function within home, hobbies) and personal care. Each is rated from 0 (unimpaired) to 3 (severely impaired). The six domains are related to the global CDR score through a complex scoring rule characterized by many if-else conditions, making it essential to identify direct dependencies. This approach will help us understand the nuanced reasons behind specific CDR scores for subjects, enabling a deeper analysis of dementia severity. The domain scores are based on independent interviews conducted by an experienced clinician and are consolidated using well-established scoring rules. To comprehend the relation between individual domain scores and the global CDR values, various models were proposed. 
For example, a study \cite{kim2017clinical} using Logistic regression analyses showed that the orientation domain in CDR is more predictive of the progression of amnestic mild cognitive impairment (aMCI) elderly subjects to Alzheimer's disease (AD) dementia. Furthermore, another study utilized item response theory (IRT) in understanding the dominant domains that influence the CDR staging and in developing a shortened version of CDR score \cite{li2021item}. 
Overall, no model in the literature explain the causal relationship between the domain scores and the global CDR, intuitively. The latent traits considered in the IRT framework are not intuitive and it is difficult to explain the dependencies between the variables. 

Causal inference models seek to learn the causal associations and the effect of interventions on outcomes. Its potential to impact healthcare research has dramatically increased in the recent times due to advancements in machine learning. Causal inference is studied broadly under two frameworks in the literature, namely the structural causal model (SCM) and the potential outcome framework (POF) \cite{shi2022learning}. One of the first steps toward computational causal inference was the development of theories connecting graphical models to causal concepts, leading to formulation of Bayesian networks \cite{spirtes2001causation}, which are represented by directed acyclic graph (DAG). DAG is a directed graph that forms a part of SCM. Its edges represent causal relationships, and the nodes represent variables including treatments, outcomes, and covariates. DAGs presume that the data satisfies the causal Markov condition, faithfulness, and causal sufficiency, which may not always be true but overall reveal patterns that convey the underpinnings in the pathways connecting input nodes to the output nodes. The advantage in studying clinical data using DAG lies in revealing dependencies which may not be apparent from the data. Further, Bayesian networks allow us to handle with incomplete data, relying on the learnt dependencies between the variables. Importantly, the effect of interventions, i.e., by changing the domain scores, on global CDR can be well-studied using DAGs \cite{heckerman1997bayesian}. 

We aim to understand the casual relationship between the global CDR and the six domain scores, along with the influences amongst the six domain scores, to simplify the complex scoring rule using Bayesian networks through DAGs formulation. We hypothesize that the DAGs not only capture the inter-dependencies between the domain scores and the global CDR but also the overall nature of the dataset under study. Therefore, we hereby employ two datasets, Alzheimer's Disease Neuroimaging Initiative (ADNI) (\textbf{\href{https://adni.loni.usc.edu} adni.loni.usc.edu}) and Longitudinal Aging Study of India (LASI) datasets, to summarize the global assessment scores and bring out the population-specific differences and similarities between these two populations. The subjects recruited as a part of ADNI project are primarily from North America, and therefore by contrasting the two datasets mentioned above, we would be able to ascertain the differences in Western and the Indian population in regards to association between the domain scores of clinical dementia rating. We hope that such an approach will help us in discovering the personality traits of the population studied under the above two projects, thereby revealing venues to focus on for reducing the chances of developing dementia at large.

\section{Related Work}
\subsection{Studies on CDR modeling}
The global cognitive assessment scores like Clinical Dementia Rating (CDR) were modelled for assessing the likeliness of a subject transitioning from healthy to mild cognitive disorder or even to Alzheimer's disease. A study \cite{ito2014predicting} employed nonlinear mixed-effect likelihood-based approach to model CDR-SB (Sum of Scores version) with subsequent in-clinic visit time. The model was fit using a population analysis approach with the Laplace Conditional Estimation method. Another study \cite{montano2013clinical} modelled dementia conversion rate using the Poisson regression method, in a cohort of Brazilian residents. Interestingly, they report that the conversion rate to dementia was higher among the subjects with CDR = 0.5 and especially higher in cases where the sum of box scores was $> 1$. On the other side, there is a study \cite{alshboul2023application} that have utilized machine learning methods to estimate the cognitive status (Cognitively Normal (CN), MCI, or dementia) of the subject, based on the sum of CDR scores (CDR-SB) besides demographic variables, such as Support Vector Machine (SVM), K-nearest neighbors (KNN), Decision Trees (C4.5), Probabilistic Naïve Bayes (NB), and Rule Induction (RIPPER). Furthermore, there was a study \cite{nakata2009combined} that a combined series of tests for orientation, memory, attention, executive function, and abstraction and judgment to discriminate subjects with MCI from healthy participants, in cases where patients live alone, without any collateral sources for overall CDR assessment score.

\subsection{Studies on utilizing DAG for modeling causal influences}
Studies on utilizing Directed Acyclic Graphs (DAGs) for modeling causal influences have shown significant advancements in understanding complex relationships within data. Pearl's foundational work on causal inference using DAGs provided a robust framework for distinguishing between correlation and causation \cite{pearl2000models}. Subsequent research has extended these concepts, applying DAGs to various domains including epidemiology, where they have been instrumental in elucidating pathways between risk factors and health outcomes \cite{hernan2020causal}. In the realm of social sciences, Morgan and Winship \cite{morgan2015counterfactuals} demonstrated the efficacy of DAGs in analyzing socio-economic data, allowing researchers to uncover hidden causal mechanisms. More recent studies, such as those by Imbens and Rubin \cite{imbens2015causal}, have integrated DAGs with machine learning techniques, enhancing the precision and interpretability of causal models. Furthermore, DAGs have been crucial in genomics for mapping gene regulatory networks, highlighting their versatility in handling high-dimensional biological data \cite{friedman2000using}. The application of DAGs in these diverse fields underscores their pivotal role in advancing causal inference methodologies.

These two distinct areas of research, are aiding each other gradually with increase in the usage of causal models in clinical research. Interestingly, usage of DAGs in the context of clinical data, was reported by one of the recent studies \cite{ding2018hybrid} harnessed Bayesian networks in capturing causal relationship between different ageing data such as age, genotype, and markers from coarse-grained neuroimaging data like the grey matter volume,  cerebrospinal fluid volume, to explain the Clinical dementia rating (CDR) and other cognitive and functional assessments. They considered Hill Climbing (HC) score-based learning technique to identify the DAG. Further, they tested the Bayesian network in classifying subjects into four-classes HC, very mild, mild, and moderate AD, with performance measures like the receiving operator characteristics curve (AUC), multi-class classification accuracy (MAC), sensitivity and specificity. There was yet another study \cite{seixas2014bayesian} that reported the utility of Bayesian networks in building clinical decision support systems (CDSS) to help in the diagnosis of dementia, AD and MCI.

\section{Materials and Methods}
\subsection{Clinical data}
The Alzheimer's Disease Neuroimaging Initiative (ADNI) is a remarkable initiative aimed at studying the progression of Alzheimer's disease (AD) using various biomarkers, which started in 2004. ADNI data is divided into four phases, each building on the previous ones and incorporating new objectives and methodologies. Clinical data was downloaded from (\textbf{\href{https://adni.loni.usc.edu} adni.loni.usc.edu}). We have considered 13,047 samples (2107 subjects) over the three phases, with 4,205 from first phase, 594 from GO phase, 5,073 from second phase, and 3,169 from third phase. Of them, 133 subjects had invalid CDR scores (-1) and were therefore discarded. In addition, 66 entries with invalid CDR scores as NaN were discarded. Participants within the age group of 55 to 90 years were considered for all the recordings.
The Longitudinal Ageing Study in India (LASI) is a project aimed at making available comprehensive and internationally comparable survey data from India, covering the entire range of topics necessary for understanding the economic, social, psychological, and health aspects of older adults and the aging process \cite{perianayagam2022cohort,bloom2021longitudinal}. The LASI is the India’s first and the largest longitudinal ageing survey study in the world, over 73,000 subjects aged 45 and above, across all states and union territories of India. It is a joint effort by the Harvard T.H. Chan School of Public Health (HSPH), the International Institute for Population Sciences (IIPS) in India, and the University of Southern California (USC). It is modeled after and harmonized with the Health and Retirement Study (HRS). Diagnostic Assessment of Dementia for the Longitudinal Aging Study in India (LASI-DAD) is an in-depth study of late-life cognition and dementia, over the older adults aged 60 and above from LASI. We downloaded the clinical dementia ratings data, from the LASI-DAD \cite{lee2020introduction} at the Gateway to Global Aging Data \cite{LasiDad}. There are 4096 entries, with clinical dementia ratings by three clinicians recorded during 2017-2020. Of these, 1568 entries had invalid CDR scores as NaN, which were excluded from the analysis. In the present study, we modeled the CDR data given by the first clinician rater. The distribution of subjects over the different grades of dementia, starting with healthy is tabulated below for both the ADNI and LASI data.


\begin{table}
	\caption{Distribution of subjects in the datasets}
	\centering
	\begin{tabular}{lll}
		\toprule
		CDR & \multicolumn{2}{|c|}{Number of subjects}\\
		\cmidrule(r){2-3}
		           & ADNI     & LASI \\
		\midrule
		0 & 4623  & 728     \\
		0.5 & 6277 & 1610      \\
		1  & 1496 & 160  \\
            2  & 363  & 25  \\
            3  & 89  & 5 \\
		\bottomrule
	\end{tabular}
	\label{tab:table}
\end{table}

\subsection{Estimating a Directed Acyclic Graph}
Considering the six domain scores and the global CDR as seven nodes, we estimated a DAG, using one of the constraint based methods, the well-known Peter-Clarke (PC) algorithm \cite{spirtes1991algorithm}. There are majorly, three ways to fit a DAG, namely: Constraint-based, Score-based and Hybrid approaches. The Peter Clarke algorithm is one of several constraint-based approaches for estimating DAGs from data, known for its robustness in handling large datasets and its precision in identifying causal relationships, while its iterative refinement process helps to continuously update the DAG. By performing conditional independence tests and methodically orienting edges, the algorithm uncovers the underlying structure of the data. Its ability to accurately infer causal connections makes it invaluable in fields such as bio-informatics \cite{roy2013causality}, network biology \cite{ni2018bayesian}, and epidemiology \cite{tennant2021use,greenland1999causal}. 
The edge strengths can be estimated by various methods \cite{heckerman1995learning,koller2009probabilistic} such as the change in scores such as BIC or AIC with edge considered and ignored, statistic for the  distributions of the nodes connected by the edge,  etc. We considered the chi-square statistic for the edge as edge weight for comparison. We utilized chi-square statistic, normalized by the maximal statistic over all the edges, to assign edge strengths (ranging from 0 to 1). These strengths aid in inferring the degree of influence of each functional domain on both the overall dementia rating (CDR) and each other.

\subsection{Notation}
The six domain scores involved in the computation of Clinical Dementia Rating are referred by their respective acronyms henceforth. Namely, Memory by \verb+M+, Orientation by \verb+O+, Judgement and problem-solving by \verb+JPS+, Community affairs by \verb+CA+, Home and Hobbies by \verb+HH+, Personal care by \verb+PC+ as referred by the main reference \cite{morris1993clinical}.

\section{Results}
The DAGs were estimated for the CDR data considered from the ADNI and LASI databases using the PC algorithm, along with the chi-square-based edge weight across the seven nodes, including six domain scores, and the global CDR. The structural learning gave rise to a relatively sparse DAG in LASI, compared with the ADNI dataset. This reduction in edges is elucidated in terms of node degrees for the two DAGs in table \ref{tab:table1}. Besides, the edge strengths of the major edges, incoming on \verb+M+ and \verb+CDR+ are shown in table \ref{tab:table2}. 

\subsection{Similarities in the Bayesian network structure of ADNI and LASI}
In terms of similarities, the edge from \verb+PC+ to \verb+CA+ has a similar causal direction and also remains the edge with the lowest edge strength among the two DAGs, while the edge between \verb+M+ and \verb+CDR+ came out to be the strongest, with the largest edge weight. Overall, it is also worth noting that in both the DAGs, all six domain scores had outgoing edges to Global, without explicit constraints imposed on the structure learning algorithm. In addition, the edges that showed similar orientation between the nodes were: \verb+CA+ to \verb+HH+, \verb+O+ to \verb+JPS+, \verb+O+ to \verb+M+, \verb+JPS+ to \verb+M+, \verb+PC+ to \verb+HH+, \verb+PC+ to \verb+CA+.

\begin{table}
	\caption{Differences in Node degrees}
	\centering
	\begin{tabular}{lllll}
	\toprule
	Node & \multicolumn{2}{|c|}{ADNI} & \multicolumn{2}{|c|}{LASI} \\
	\cmidrule(r){2-3}
        \cmidrule(r){4-5}
	      & Incoming  & Outgoing & Incoming  & Outgoing \\
	\midrule
	\verb+CDR+ & 6  & 0 & 6 & 0     \\
	\verb+M+     & 4 & 1 & 2 & 1      \\
	\verb+PC+    & 1  & 2 & 0 & 3  \\
        \verb+CA+  & 1  & 5   & 2  & 2  \\
        \verb+O+  & 4  & 1   & 0  & 3  \\
        \verb+JPS+    & 0 & 5  & 1 & 3 \\
        \verb+HH+    & 3 & 3 & 2 & 1  \\
		\bottomrule
	\end{tabular}
	\label{tab:table1}
\end{table}

\subsection{Differences in the ADNI and LASI DAGs}
Of all the differences between the DAGs, strikingly, \verb+HH+ and \verb+CA+ has no influence on \verb+M+ in LASI DAG, but they had an influence in ADNI DAG. Interestingly, the edge between the \verb+JPS+ and \verb+CA+ reversed in direction. The following edges disappeared in LASI over ADNI, namely: \verb+CA+ to \verb+M+, \verb+CA+ to \verb+JPS+, \verb+O+ to \verb+CA+, \verb+HH+ to \verb+M+, \verb+HH+ to \verb+O+, and \verb+JPS+ to \verb+HH+. While there is a similar dominance in causal influence of \verb+O+ on \verb+M+ in LASI and ADNI DAGs, with 0.8 edge strength. Furthermore, \verb+PC+ shows the minimum influence on  \verb+CDR+ in both datasets, with 0.6 and 0.2 in ADNI and LASI DAGs respectively. The edge strengths estimated for the ADNI and LASI datasets varied considerably for a few important edges such as those between \verb+CDR+ and \verb+PC+, and those incoming on the \verb+M+.

\begin{table}
	\caption{Differences in edge strengths}
	\centering
	\begin{tabular}{llll}
		\toprule
		\multicolumn{2}{|c|}{Edge strength} & ADNI  & LASI \\
		\cmidrule(r){1-2}
		Source & Sink \\
		\midrule
		\verb+M+ & \verb+CDR+  & 1.0 & 1.0   \\
		\verb+CA+ & \verb+CDR+   & 0.9 & 0.6 \\
		\verb+HH+ & \verb+CDR+    & 0.9  & 0.8 \\
            \verb+O+ & \verb+CDR+  & 0.9  & 0.9  \\
            \verb+PC+ & \verb+CDR+  & 0.6  & 0.2 \\
            \verb+JPS+ & \verb+CDR+  & 0.9 & 0.6  \\
            \verb+CA+ & \verb+M+   & 0.7 & - \\
		\verb+HH+ & \verb+M+    & 0.7  & - \\
            \verb+O+ & \verb+M+  & 0.8  & 0.8  \\
            \verb+JPS+ & \verb+M+  & 0.8 & 0.5  \\
		\bottomrule
	\end{tabular}
	\label{tab:table2}
\end{table}

\begin{figure*}[!t]
\centering
\subfloat[]{\includegraphics[width=3in]{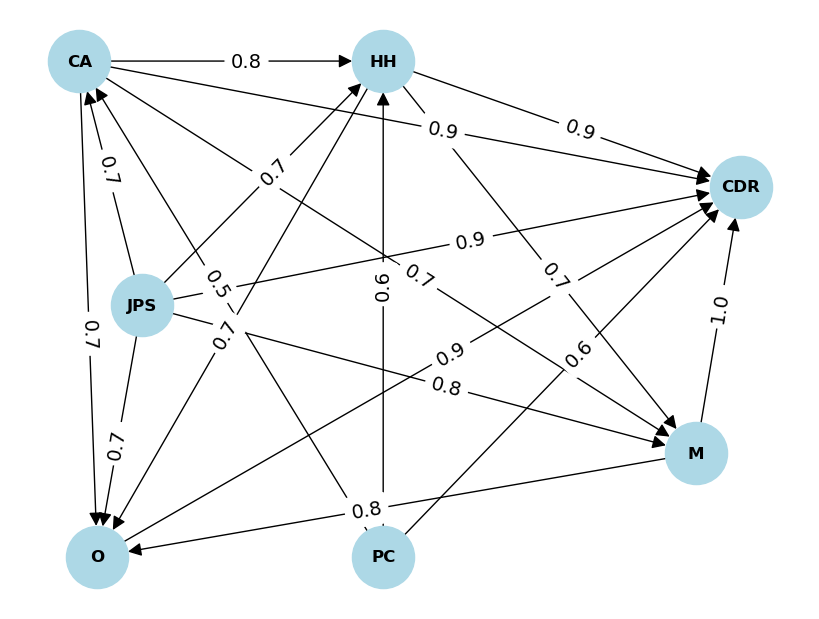}}%
\label{fig_first_case}
\hfil
\subfloat[]{\includegraphics[width=3in]{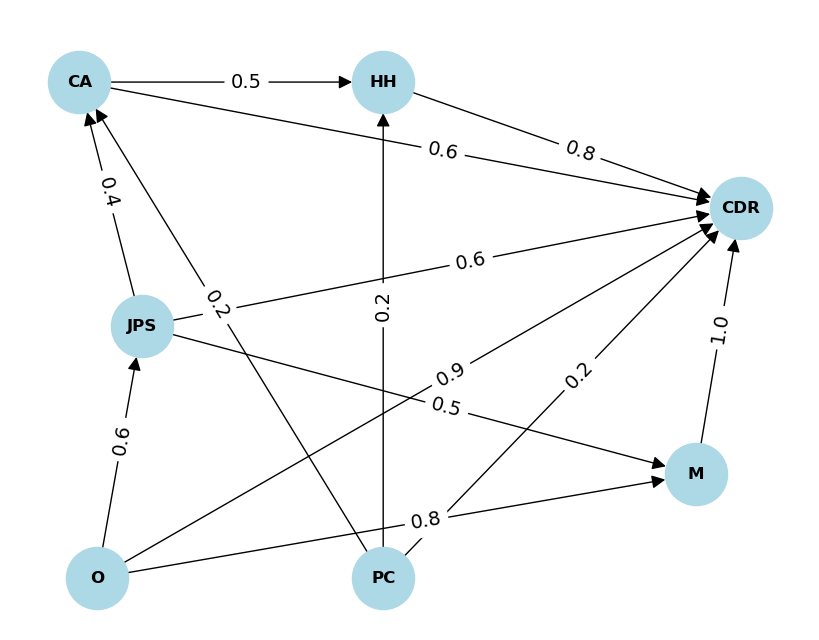}}%
\label{fig_second_case}
\caption{Reconstructed DAG structure for the two datasets, using PC algorithm. (a) ADNI dataset. (b) LASI dataset. The full form of the six cognitive domains are: \texttt{M} - Memory, \texttt{O} - Orientation, \texttt{JPS} - Judgement and problem-solving, \texttt{CA} - Community affairs, \texttt{HH} - Home and Hobbies, \texttt{PC} - Personal care}
\label{fig_sim}
\end{figure*}

\section{Discussions}

\subsection{Differences in ADNI and LASI datasets}
The ADNI dataset considered for our study here was from the 4 phases namely, ADNI1, ADNI2, ADNI3 along with ADNIGO phase. Most of the subjects in the cohorts, previous to the ADNI4 phase were mainly recruited from the United States and Canada \cite{alzforum_adni}. Specifically, the subjects were either white or non-Hispanic. Therefore, the representation of other ethnic and racial groups may be limited compared to the majority Western population within the study. There have been attempts in the recent times to include diverse population in terms of their descent, medical history, age, and sex. ADNI4 has a focused goal of recruiting at least 50\% of its new participants from diverse and underrepresented populations.
While the LASI (Longitudinal Ageing Study in India) dataset was recorded from Indians, who were primarily from the Indian subcontinent, it provides a valuable perspective on aging and dementia in a Indian context. The LASI dataset includes a diverse cohort from various regions within India, capturing a range of socio-economic backgrounds and ethnic groups. This diversity helps address the gaps in representation found in studies like ADNI, offering insights into the impact of cultural and regional factors on aging and cognitive health.

\subsection{Contrasting DAG with other models on CDR}
The Item Response Theory model was employed in delineating the dependence of domain scores on the global CDR \cite{li2021item}. They reported that the Memory and Orientation domains are most-correlated in terms of the factors estimated from the IRT model. Interestingly, most of the questions (6/11) in  Community Affairs domain, turned out to convey minimal information about the global CDR. When considering bi-factor IRT model with correlated domain specific factors, factor loading for each questionnaire, showed striking pattern. Based on the general factor loadings, the study partner questions had much larger value than the study participant questions. Whereas, it was opposite with the study participant questions having larger value, based on domain specific factor loadings. The model-derived IRT scores provide a continuous measure of dementia severity.

Bayesian network models, represented by Directed Acyclic Graphs (DAGs), and Item Response Theory (IRT) models, used to understand the importance of each domain score in determining the global Clinical Dementia Rating (CDR), serve different purposes and are applied in distinct contexts. Primarily, IRT focuses on the measurement and scaling of latent traits, whereas Bayesian networks concentrate on causal inference and probabilistic reasoning. Nevertheless, both methods convey valuable insights into the structure and relationships within complex datasets. IRT models are generally employed to analyze psychometric tests and questionnaires, modeling the relationship between latent traits and item responses using probabilistic functions. These models facilitate the understanding of how different test items reflect underlying latent traits, thus enhancing the accuracy of assessments and interpretations in psychological and educational research.

\subsection{Advantages of PC algorithm, especially for the categorical data like CDR data}
PC algorithm is one of the global discovery algorithms in the category of constraint-based approaches which aim to learn the entire graph structure as a whole \cite{Spirtes_2000, Kalisch_2012}, unlike local constraint-based algorithms that focus on the local structure related to each variable independently. 
Global discovery algorithms, such as the PC algorithm, offer the advantage of learning the entire network structure simultaneously, which allows for a comprehensive understanding of variable interdependencies \cite{Spirtes_2000}. Unlike score-based methods, these algorithms do not require a global scoring function, making them more suitable for categorical data and complex datasets \cite{Tsamardinos_2006}. This holistic approach facilitates the identification of intricate relationships between variables, which is crucial for accurate modeling in fields like clinical research \cite{Kalisch_2012}. Constraint-based methods unlike score-based, do not require a global scoring function which assess the goodness of fit of the network structure over the entire dataset. Interestingly, in the case of categorical data like with CDR data, these can be hard to define and optimize \cite{Tsamardinos_2006, Neapolitan_2004}. Interestingly, the relationships in clinical datasets are often localized \cite{Murphy_2012, Chickering_2002}, meaning that variables directly related to a particular clinical measure are more likely to be interdependent. Therefore, by focusing on adjacent nodes, the PC algorithm can effectively uncover the localized dependencies.

\subsection{Field testing}
Field testing is crucial to validate the causal inferences made through DAGs, as clinicians' insights can provide context and interpretation beyond the statistical model. For instance, expert input can help refine the edge weights and validate the clinical relevance of the inferred relationships, addressing discrepancies between model predictions and clinical observations \cite{pearl2000models, Spirtes_2000}. Without such clinical validation, the DAGs may not fully capture the complexity of the disease process or the nuances of patient data \cite{heckerman1995learning}. Engaging clinicians ensures that the models are not only theoretically sound but also practically applicable in real-world settings \cite{Schölkopf2021}

\section{Conclusion}
The Clinical Dementia Rating score is determined by the clinician, based on the questionnaire asked under the six domains. The process of estimating the score based on the rating for each of the six domains is rigorous, and cumbersome. We hereby, came up with a causal model using DAGs to understand the association between the six domain scores and global CDR score. The objective was to study the dependence of causal structure on the subject population used for reconstruction. We tested the data recorded as part of ADNI and LASI projects, which sampled varied regions from North America, and India respectively. We found that the LASI dataset resulted in a sparse DAG with diverse edge weights, while ADNI DAG was relatively dense with narrow distribution of edge weights around high values. Such differences in Bayesian network structure can well inform the clinicians of the importance of the six cognitive domains in decision making. Striking similarity across the two DAGs was that both had high influence of memory domain on global score, and among those incoming on memory, orientation dominated. Furthermore, among the edges incoming on memory, domains such as Home and hobbies, and Community affairs had relatively lower edge strengths in ADNI DAG, while they disappeared in LASI DAG. The present study highlights the utility of causal models such as Bayesian networks in studying the dependency of various domains on CDR scores, for strategizing therapies to address dementia at large.



\bibliographystyle{IEEEtran}
\bibliography{MAIN}


 


\vspace{20pt}
\section{Biography Section}
\vspace{-15pt}
\begin{IEEEbiographynophoto}{Wupadrasta Santosh Kumar}
 received the Ph.D. degrees in Systems Neuroscience from Indian Institute of Science (IISc), in 2023. He is currently a Post Doctoral Fellow candidate at the Centre for Brain Research (CBR). His research interests include EEG signal processing, graph theoretic approaches to study multi-channel time series, and causal inference.
\end{IEEEbiographynophoto}

\vspace{-20pt}
\begin{IEEEbiographynophoto}{Sayali Rajendra Bhutare}
 completed her Bachelor’s degree in Biotechnology from Pune university. She is currently pursuing her Masters degree in life Sciences at the Indian Institute of Science (IISc) majoring neuroscience and behaviour. She was a research intern at the Centre for Brain research (CBR) in summer 2024. She is interested in understanding the neural mechanisms in the brain that underpin cognitive and behavioral tasks, as well as studying brain diseases and disorders.
\end{IEEEbiographynophoto}

\vspace{-20pt}
\begin{IEEEbiographynophoto}{Neelam Sinha}
 is a faculty at Center for Brain Research, an autonomous center at IISc, Bangalore. Her interests lie in applying machine learning techniques for multi-modal neuroimaging. Before joining CBR, she was a faculty at IIIT-Bangalore. At IIIT-Bangalore, her research focus was on problems in healthcare, which included surgical video, fundal image and neuro-data analysis, in collaboration with NIMHANS, Bangalore. 
\end{IEEEbiographynophoto}

\vspace{-20pt}
\begin{IEEEbiographynophoto}{Thomas Gregor Issac}
 completed his M.B.B.S training (2009) at M.O.S.C Medical College (Mahatma Gandhi University), Kerala. Subsequently, he was selected for the Ph.D. in Clinical Neurosciences Programme through the ICMR MD-PhD Talent Search programme. During his Ph.D., he investigated the genetic, cognitive, and neuroimaging profiles of a cohort comprising over 200 patients across the spectrum of vascular cognitive impairment. Following this, he pursued the MD Psychiatry course at NIMHANS and served as a Junior Resident in the Department of Psychiatry from July 2016 to June 2019. During his MD, he focused on studying the role of the APO E4 allele in vascular cognitive impairment. Additionally, he successfully cleared the DNB in Psychiatry. Subsequently, he completed his DM degree in Geriatric Psychiatry from NIMHANS. Presently, his research lab is dedicated to clinical research, with a focus on neurocognition, neuropsychiatric disorders, geriatric issues, neuro interventions, geriatric advocacy, and social policy.
\end{IEEEbiographynophoto}
\vfill

\end{document}